\documentclass[10pt, a4paper]{article}
\usepackage{lrec}
\usepackage{graphicx}
\usepackage{tabularx}
\usepackage{soul}

\usepackage{epstopdf}

\usepackage{hyperref}
\usepackage{xstring}

\usepackage{multirow}
\usepackage{makecell}
\usepackage{amsmath,amssymb,amsfonts}
\usepackage[figuresright]{rotating}  
\usepackage[table]{xcolor}
\usepackage[draft=true]{minted}  

\usepackage{CJKutf8}

\newcommand{\indep}{\;\rotatebox[origin=c]{90}{$\models$}\;}

\newcommand{\korean}[1]{\begin{CJK*}{UTF8}{mj}{#1}\end{CJK*}}
\newcommand{\traditionalchinese}[1]{\begin{CJK*}{UTF8}{bsmi}{#1}\end{CJK*}}

\newcommand{\zh}[1]{\begin{CJK*}{UTF8}{gbsn}{#1}\end{CJK*}}

\title{\emph{word2word}: A Collection of Bilingual Lexicons for 3,564 Language Pairs}

\name{Yo Joong Choe\textsuperscript{*}\thanks{\textsuperscript{*}Equal contribution.}, Kyubyong Park\textsuperscript{*}, Dongwoo Kim\textsuperscript{*}}

\address{Kakao Brain \\
         20, Pangyoyeok-ro 241, Bundang-gu, Seongnam-si, Gyeonggi-do, Korea \\
         \texttt{\{yj.choe,kyubyong.park,dongwoo.kim\}@kakaobrain.com}\\}

\abstract{
  We present \emph{word2word}, a publicly available dataset and an open-source Python package for cross-lingual word translations extracted from sentence-level parallel corpora. 
  Our dataset provides top-$k$ word translations in 3,564 (directed) language pairs across 62 languages in OpenSubtitles2018 \cite{lison2018opensubtitles2018}.
  To obtain this dataset, we use a count-based bilingual lexicon extraction model based on the observation that not only source and target words but also source words themselves can be highly correlated.
  We illustrate that the resulting bilingual lexicons have high coverage and attain competitive translation quality for several language pairs. 
  We wrap our dataset and model in an easy-to-use Python library, which supports downloading and retrieving top-$k$ word translations in any of the supported language pairs as well as computing top-$k$ word translations for custom parallel corpora.
   \newline  
}

\begin{document}

\maketitleabstract

\section{Introduction}

Bilingual lexicons \cite{fung1998statistical} are valuable resources for cross-lingual tasks, including low-resource machine translation \cite{ramesh2018neural,gu2019pointer} and cross-lingual word embeddings \cite{ruder2017survey}.
However, it is often difficult to find a large enough set of bilingual lexicons that is freely and readily available across various language pairs \cite{levy2017strong}.
For example, standard bilingual dictionaries like Wiktionary\footnote{\scriptsize\url{https://en.wiktionary.org}} often do not explicitly provide word correspondences but refers or redirects to the query word's dictionary form:
\begin{itemize}
    \item {\bf Query:} \emph{travaill\'e} (French for {`worked'}) \\
    {\bf Result:} (verb) past principle of \emph{travailler} {`work'}
    \item {\bf Query:} {\korean{먹었다}} (Korean for {`ate'}) \\
    {\bf Result:} \emph{redirects to \emph{\korean{먹다}} {`eat'}}
\end{itemize}
Not only does this make it tedious to find word-level correspondences across many query words, this is particularly problematic when we try to find word correspondences for languages where some dictionary forms are rarely used in ordinary discourse, such as the case of \korean{먹다} in the Korean language. 

While the task of bilingual lexicon extraction (BLE) has been popular in both early and recent literature, spanning from count-based approaches \cite{fung1998statistical,vulic2013cross,liu2013topic} to using cross-lingual word embeddings \cite{ruder2017survey,mikolov2013exploiting,gouws2015bilbowa,conneau2017word,levy2017strong,artetxe2018robust,artetxe2019bilingual}, few were focused on building high-coverage bilingual lexicons across many language pairs, possibly including non-Indo-European languages.
In fact, many of the recent studies and their accompanying packages \cite{conneau2017word,artetxe2018robust,glavas2019properly} aim at evaluating cross-lingual word embeddings, so that they involve at most 10-100s of language pairs and 1-5K words for each pair.

Motivated by the lack of publicly available and high-coverage bilingual lexicons across diverse languages, we present \emph{word2word}, a large collection of bilingual lexicons for 3,564 language pairs across 62 languages that is wrapped around an open-source and easy-to-use Python interface. 
We extract top-$k$ bilingual word correspondences from all parallel corpora provided by OpenSubtitles2018\footnote{\scriptsize{\url{http://opus.nlpl.eu/OpenSubtitles-v2018.php}}} \cite{lison2018opensubtitles2018}, using a count-based model that takes into account both monolingual and cross-lingual co-occurrences.
The package also provides interface for obtaining bilingual lexicons for custom parallel corpora in any other language pairs and domains not covered by OpenSubtitles2018. 

\section{The \emph{word2word} Dataset}

\begin{table}[t]
\centering
\begin{tabular}{c | c}
\Xhline{1.1pt}
\# Languages & 62 \\
\# Language Pairs & 3,564 \\
Avg. Lexicon Size & 127,023 \\
Avg. \# Translations Per Word & 8.8 \\
\Xhline{1.1pt}
\end{tabular}
\caption{Overview of the \emph{word2word} dataset.}
\label{tbl:overview}
\end{table}

\begin{table*}[ht]
\centering
\small
\rowcolors{1}{white}{gray!15}
\begin{tabular}{l|r|r|r|r}
\Xhline{1.1pt}
\bf Language Pair    & \bf Lexicon Size & \bf \# Unique Translations & \bf Avg. \# Translations Per Word & \bf \# Sentences Used \\ \hline
Arabic-English       &           335.5K &                 86.0K &                      9.7 &                 29.8M \\
English-Arabic       &            97.6K &                191.6K &                      9.5 &                 29.8M \\
S.Chinese-English    &           214.0K &                 87.0K &                      9.5 &                 11.2M \\
English-S.Chinese    &           101.6K &                139.1K &                      9.4 &                 11.2M \\
T.Chinese-English    &           201.7K &                 72.5K &                      9.5 &                  4.8M \\
English-T.Chinese    &            85.8K &                119.7K &                      9.2 &                  4.8M \\
French-English       &            92.1K &                 59.1K &                      9.8 &                 41.8M \\
English-French       &            72.1K &                 71.4K &                      9.7 &                 41.8M \\
Italian-English      &           111.5K &                 63.9K &                      9.7 &                 35.2M \\
German-English       &           127.0K &                 64.8K &                      9.7 &                 22.5M \\
English-German       &            73.6K &                 95.9K &                      9.6 &                 22.5M \\
English-Italian      &            75.4K &                 83.9K &                      9.6 &                 35.2M \\
Japanese-English     &            83.3K &                 75.2K &                      9.2 &                  2.1M \\
English-Japanese     &           102.1K &                 63.8K &                      9.3 &                  2.1M \\
Korean-English       &            87.2K &                 75.8K &                      9.3 &                  1.4M \\
English-Korean       &           105.5K &                 69.8K &                      9.1 &                  1.4M \\
Russian-English      &           213.4K &                 68.7K &                      9.7 &                 25.9M \\
English-Russian      &            76.2K &                155.8K &                      9.5 &                 25.9M \\
Spanish-English      &           107.1K &                 60.8K &                      9.8 &                 61.4M \\
English-Spanish      &            73.9K &                 82.5K &                      9.7 &                 61.4M \\
Thai-English         &           155.6K &                 84.2K &                      9.4 &                  3.3M \\
English-Thai         &           109.2K &                 99.2K &                      9.2 &                  3.3M \\
Vietnamese-English   &            96.6K &                 76.6K &                      9.0 &                  3.5M \\
English-Vietnamese   &            96.4K &                 75.3K &                      9.3 &                  3.5M \\
\Xhline{1.1pt}
\end{tabular}
\caption{Summary statistics for the \emph{word2word} dataset between selected languages and English. 
Lexicon size refers to the number of unique words in source language for which translations exist. 
S.Chinese and T.Chinese refer to simplified and traditional Chinese, respectively.
}
\label{tbl:summary}
\end{table*}

\setlength\tabcolsep{2pt}
\begin{table*}[t]
\centering
\begin{tabular}{c | c c c c c}
\Xhline{1.1pt}
{\bf Word} & \multicolumn{5}{c}{{\bf Top-5 Translations}} \\
\hline
    \rowcolor{gray!15} {English} & \multicolumn{5}{c}{{French}} \\ \hline
     exceptional & exceptionnel & exceptionnelle & exceptionnels & exceptionnelles & exception \\
     whether & plaise & décider & importe & question & savoir \\
     committee & comité & éthique & accueil & commission & central \\
     clown & clown & clowns & bouffon & guignol & cirque \\
     spread & dispersez-vous & propagation & répandre & propager & répandu \\
    \hline
    \rowcolor{gray!15} {French} & \multicolumn{5}{c}{{English}} \\ \hline
     hobbs & hobbs & abigail & garret & jacob & garrett \\
     mêlé & mixed & involved & middle & part & murder \\
     établir & establish & establishing & set & able & connection \\
     taule & slammer & joint & locked & jail & prison \\
     chaussettes & socks & sock & stockings & pairs & underwear \\
\hline
    \rowcolor{gray!15} { English} & \multicolumn{5}{c}{{ Korean}} \\ \hline
     slaughtered & \korean{학살} & \korean{도륙} & \korean{도살} & \korean{당했} & \korean{살육} \\
     shadow & \korean{그림자} & \korean{그늘} & \korean{알맞} & \korean{어둠} & \korean{존재} \\
     Charles & \korean{찰스} & \korean{제프리} & \korean{Charles} & \korean{조프리} & \korean{램퍼트} \\
     concerns & \korean{걱정} & \korean{우려} & \korean{염려} & \korean{관한} & \korean{판단력} \\
     reverse & \korean{역} & \korean{뒤집} & \korean{후진} & \korean{거꾸로} & \korean{되돌리} \\
    \hline
    \rowcolor{gray!15} {Korean} & \multicolumn{5}{c}{{English}} \\ \hline
     \korean{아유} & arm & Thank & thrilled & killing & NamWon \\
     \korean{상어} & shark & Shark & sharks & Tank & Tiger \\
     \korean{쥐} & rat & rats & mouse & mice & squeeze \\
     \korean{기꺼이} & willing & happy & pleasure & gladly & willingly \\
     \korean{어떤} & Some & kind & which & any & anything \\
\Xhline{1.1pt}
\end{tabular}
\caption{Randomly sampled words and their top-5 translations in the English$\leftrightarrow$French and English$\leftrightarrow$Korean \emph{word2word} bilingual lexicons. Top-5 translations are listed in the descending order of scores.}
\label{tbl:examples}
\end{table*}

\subsection{Data Statistics}

The \emph{word2word} dataset spans across 3,564 directed language pairs between 62 languages in the OpenSubtitles2018 dataset, a collection of translated movie subtitles extracted from OpenSubtitles.org\footnote{\scriptsize{\url{http://www.opensubtitles.org/}}}.
By design, our methodology covers 100\% of words present in the source sentences, making the lexicon size much larger than existing bilingual dictionaries. 
The lexicon also contains up to top-10 word translations in the target language.
We provide an overview of the entire dataset in Table \ref{tbl:overview}.

In Table \ref{tbl:summary}, we provide summary statistics for bilingual lexicons between English and some of the major languages (both European and non-European).
For each pair, the lexicon size ranges from 76.2K (English-Russian) to 335.5K (Arabic-English), demonstrating the broad coverage of words in the dataset.
For each of these words, the dataset includes an average of 9 or more highest-scored translations according to our extraction approach described in Section \ref{sec:ble}
Lexicon size for all language pairs can be found in Appendix \ref{app:counts}.

\subsection{Examples}

In Table \ref{tbl:examples}, we present samples of top-5 word translations in the English$\leftrightarrow$French and English$\leftrightarrow$Korean bilingual lexicons.
For each language pair, we randomly sample five words from the top-10,000 frequent words in the source lexicon and provide their top-5 word translations.
This is to show translations for words that are relatively more likely used than others in typical discourse.

\section{Methodology}

\subsection{Bilingual Lexicon Extraction}\label{sec:ble}

Bilingual lexicon extraction (BLE) is a classical natural language task where the goal is to find word-level correspondences from a (parallel) corpus.
There are many different approaches to BLE, such as word alignment methods \cite{brown1993mathematics,vogel1996hmm,koehn2007moses} and cross-lingual word representations \cite{ruder2017survey,mikolov2013exploiting,liu2013topic,gouws2015bilbowa,conneau2017word}.

Among them, we focus on simple approaches that can work well with various sizes of parallel corpora that are present in OpenSubtitles2018, which ranges from 129 sentence pairs in Armenian-Indonesian to 61M sentence pairs in English-Spanish. 
In particular, we avoid methods that require high-resource parallel corpora (e.g., neural machine translation) or external corpora (e.g., unsupervised or semi-supervised cross-lingual word embeddings).
Also, since bilingual word-to-word mappings are hardly one-to-one \cite{fung1998statistical,somers2001bilingual,levy2017strong}, we consider methods that yield relevance scores between every source-target word pair, such that we can extract not just one but the top-$k$ correspondences.
For these reasons, we consider approaches based on (monolingual and cross-lingual) co-occurrence counts: co-occurrences, pointwise mutual information (PMI), and co-occurrences with controlled predictive effects (CPE).

\subsubsection{Co-occurrences}\label{sec:cooccurrences}

The simplest baseline for our goal is to compute the co-occurrences between each source word $x$ and target word $y$. 
For each source word $x$, we can score any target word $y$ based on the conditional probability $p(y|x) \propto p(x, y)$:
\begin{align}
p(y | x) &= \frac{p(x, y)}{p(x)} \approx \frac{\#(x, y)}{\#(x)} \propto \#(x, y) \label{eqn:condcount}
\end{align}
where $\#(\cdot)$ denotes the number of (co-)occurrence counts of the word or word pair across the parallel corpus.
The top-$k$ translations of source word $x$ can be computed as the top-$k$ target words with respect to their co-occurrence counts with $x$.

\subsubsection{Pointwise Mutual Information}\label{sec:pmi}

Another simple baseline is pointwise mutual information (PMI), which further accounts for the monolingual frequency of a candidate target word $y$:
\begin{align}
\mathsf{PMI}(x, y) &= \log \frac{p(x, y)}{p(x)p(y)} \\ \nonumber
&\approx \log \frac{\#(x, y)}{\#(x)\#(y)} \propto \log\#(x, y) - \log\#(y) \label{eqn:pmicount}
\end{align}
Compared to the co-occurrence model in \eqref{eqn:condcount}, PMI can help prevent stop words from obtaining high scores.

The use of PMI has been connected to the skip-gram with negative sampling (SGNS) \cite{levy2014neural} model of \emph{word2vec} \cite{mikolov2013distributed}.
PMI can also be interpreted as a conditional version of TF-IDF \cite{fung1998statistical}.

\subsubsection{Controlled Predictive Effects}\label{sec:cpe}

\begin{figure}
    \centering
    \includegraphics[height=6.5cm,clip=false]{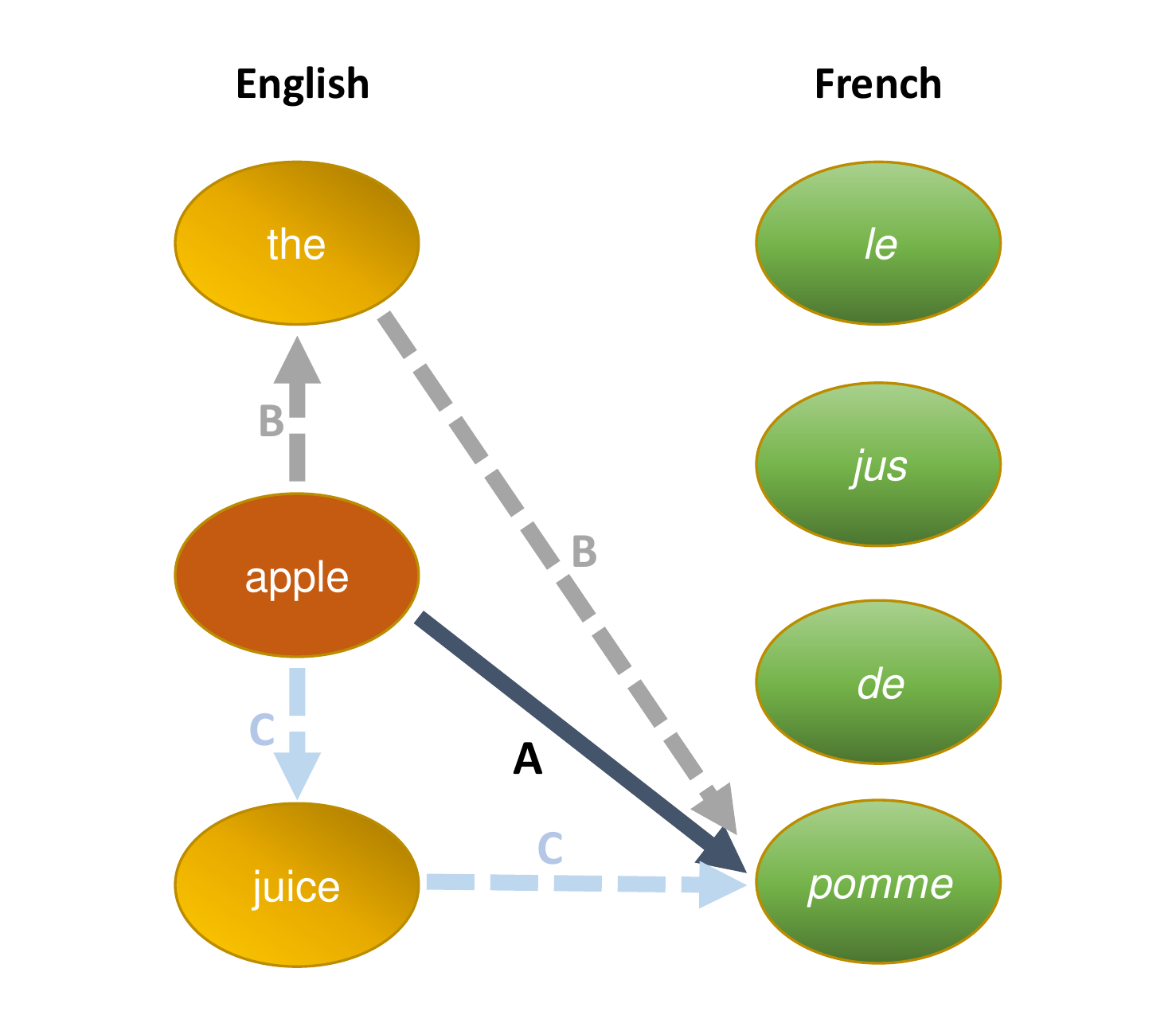}
    \caption{A schematic graphical model of English and French words. Co-occurrence and PMI models focus on the relationship from \emph{apple} to \emph{pomme} ({\bf A}). CPE further controls for the confounding effect of other collocates like \emph{the} (B) and \emph{juice} (C).}
    \label{fig:apple}
\end{figure}

\setlength\tabcolsep{2pt}
\begin{table*}[t]
    \centering
    \begin{tabular}{c | c | c c | c c | c c | c c | c c | c c}
    \Xhline{1.1pt}
    {\bf Metric (\%)} & {\bf Method} & {\bf en-es} & {\bf es-en} & {\bf en-fr} & {\bf fr-en} & {\bf en-de} & {\bf de-en} & {\bf en-ru} & {\bf ru-en} & {\bf en-zh} & {\bf zh-en} & {\bf en-it} & {\bf it-en} \\ \hline    \multicolumn{2}{c|}{\# Sentence Pairs} & \multicolumn{2}{c|}{61.4M} & \multicolumn{2}{c|}{41.8M} & \multicolumn{2}{c|}{22.5M} & \multicolumn{2}{c|}{25.9M} & \multicolumn{2}{c|}{4.8M} & \multicolumn{2}{c}{35.2M}  \\ \hline
    \multirow{4}{*}{P@1} & Co-occurrence &  22.3 & 25.5 & 18.7 & 21.9 & 10.5 & 23.5 & 3.3 & 11.4 & 5.4 & 3.8 & 24.9 & 24.1 \\
                               & PMI & 72.7 & 72.3 & 73.9 & 72.1 & 62.1 & 71.9 & 32.8 & 55.0 & 24.8 & 33.1 & 68.1 & 69.5  \\
                               & MUSE & 81.7 & {\bf 83.3} & 82.3 & {\bf 82.4} & 74.0 & 72.4 & 51.7 & 63.7 & 42.7 & 37.5 & 66.2 & 58.7 \\
                               & CPE & {\bf 82.4} & 79.5 & {\bf 83.6} & 80.7 & {\bf 82.4} & {\bf 81.1} & {\bf 66.7} & {\bf 68.9} & {\bf 56.0} & {\bf 58.7} & {\bf 80.9} & {\bf 82.1}  \\ \hline
    \multirow{4}{*}{P@5} & Co-occurrence & 67.8 & 71.4 & 63.1 & 66.3 & 63.7 & 65.5 & 52.3 & 51.8 & 46.0 & 36.3 & 61.9 & 68.5 \\ 
                                 & PMI & {\bf 92.3} & {\bf 90.4} & {\bf 92.5} & {\bf 90.1} & 90.5 & {\bf 88.1} & 74.1 & 79.5 & 58.7 & 66.1 & {\bf 90.3} & {\bf 91.1}
   \\
                                 & MUSE & - & - & - & - & - & - & - & - & - & - & 80.4 & 76.5  \\
                                 & CPE & 90.1 & 88.4 & 91.7 & 89.3 & {\bf 90.7} & 87.7 & {\bf 79.5} & {\bf 80.0} & {\bf 73.5} & {\bf 72.8} & 89.8 & 89.9
 \\
                            \hline
    \Xhline{1.1pt}
    \end{tabular}
    \caption{Precision (\%) on 1,500 word translations (test split from MUSE) for language pairs evaluated in the MUSE paper. 
    P@1 and P@5 denote the precision of top-1 and top-5 predictions, respectively. 
    The ISO 639-1 language codes are used (en: English, es: Spanish, fr: French, de: German, ru: Russian, zh: traditional Chinese, it: Italian).}
    \label{tbl:results_muse}
\end{table*}

While conditional probability and PMI are proportional to cross-lingual co-occurrence counts, they can fail to distinguish exactly which source word in the sentence is the most predictive of the corresponding target word in the translated sentence. 
For example, given a English-French pair (\emph{the apple juice},  \emph{la jus de pomme}), these baseline methods cannot isolate the effect of \emph{apple}, as opposed to \emph{the} and \emph{juice}, on \emph{pomme}. 

To deal with this issue, we add a correction term that averages the probability of seeing $y$ given a confounder $x'$ in the source language, i.e. $p(y|x')$. 
This probability is then weighted by the probability of actually seeing that confounder, i.e. $p(x'|x)$.
This correction can be explained intuitively by the dashed arrows in the schematic graphical model in Figure \ref{fig:apple}-- it reflects the conditional independence relationships between words that the baseline models do not. 
We call the resulting approach as the method of \emph{controlled predictive effects (CPE)}.

Formally, we define the corrected CPE score as follows:
\begin{align}
\mathsf{CPE}(y \mid x) &= p(y\mid x) - \sum_{x' \in \mathcal{X}} p(y \mid x') p(x' \mid x) \nonumber \\
&= \sum_{x' \in \mathcal{X}} \mathsf{CPE}_{y \mid x} (x') p(x' \mid x)
\end{align}
where $\mathcal{X}$ is the source vocabulary and $\mathsf{CPE}_{y\mid x} (x')$ denotes the CPE term of any other source word $x'$ when predicting $y$ from $x$. 
Formally, this term is defined as
\begin{align}
\mathsf{CPE}_{y \mid x} (x') = p(y \mid x, x') - p(y \mid x')
\end{align}
This CPE term measures the effect of \emph{additionally} seeing $x$ (\emph{apple}) when predicting $y$ (\emph{pomme}), after controlling for the effect of any other $x'$ (\emph{the}), which the model views as a confounder. 
If $\mathsf{CPE}_{y \mid x} (x') = 0$, then $x \indep y \mid x'$, meaning that after observing a confounder $x'$, $x$ is no longer related to $y$.
The CPE term for each confounder $x'$ is then marginalized over all possible confounders to give a final score, weighted by the probability of seeing that confounder in a sentence with $x$. 
Note that $\mathsf{CPE}_{y \mid x} (x) = 0$, meaning that, after seeing $x$ when predicting $y$, there is no additional effect by seeing $x$ (again). 

In practice, summing the CPE scores over all words in the source vocabulary can be inefficient. 
Because many of the (unrelated) words in the vocabulary do not play a role in the confounding, we select the top-$m$ source words with the highest co-occurrence counts and correct for their effects only. 
We used $m=5,000$ in our experiments and found that using a larger $m$ did not make a meaningful difference on the quality of top-1 and top-5 correspondences.

\subsubsection{Evaluation on MUSE Bilingual Dictionaries}\label{sec:evaluation}

We first evaluate the methods on the same ground-truth bilingual dictionaries as MUSE\footnote{\scriptsize\url{https://github.com/facebookresearch/MUSE}}, a cross-lingual neural embedding model.
Each dictionary contains 1,500 words and their translations obtained using an internal translation tool from the authors. 
Although we consider MUSE's performance as a reference, we do note that it is difficult to make a fair comparison against MUSE: the count-based methods use parallel corpora from OpenSubtitles2018, while MUSE embeddings are instead learned from monolingual Wikipedia data (for its unsupervised version) and an additional 5,000-word bilingual lexicon (for its supervised version). 

In Table \ref{tbl:results_muse}, we report the top-1 and top-5 precision scores (P@1 and P@5, respectively) of the count-based methods and MUSE embeddings across twelve\footnote{The MUSE paper also presents the results on English-Esperanto and Esperanto-English, but the ground-truth dictionary is no longer available online. See \scriptsize{\url{https://github.com/facebookresearch/MUSE/issues/34}}.} directed language pairs that were used to evaluate MUSE in its paper \cite{conneau2017word}: English-Spanish, English-French, German-English, English-Russian, English-Chinese (traditional), and English-Italian, all in both directions. 
For MUSE, we report its best reported performance (only top-1 precision is reported, except for en-it and it-en) among its supervised and unsupervised variants. 

Our main finding is that the CPE method consistently and significantly outperforms the co-occurrence and PMI baselines at top-1 precision score. 
We also find that CPE outperforms MUSE on most of the reported language pairs, especially when the number of sentence pairs is comparatively small (e.g., 13-21\% improvement between English and Chinese, for which there are about 6\% as many sentence pairs as those between English and Spanish).
In terms of the top-5 precision score, the CPE method performs comparatively well with the PMI method, which performs better on some of the selected language pairs. 
Compared to the CPE method, we suspect that the PMI method overly favors rare words because it directly penalizes word counts, so that the most likely correspondence (which isn't necessarily the least common) is pushed back to later ranks. More examples can be found in Appendix \ref{app:comparison}

\setlength\tabcolsep{2pt}
\begin{table*}[t]
    \centering
    \begin{tabular}{c | c | c c |  c c | c c | c c | c c | c c}
    \Xhline{1.1pt}
    {\bf Metric (\%)} & {\bf Method} & {\bf en-ar} & {\bf ar-en} & {\bf en-zh} & {\bf zh-en} & {\bf en-ja} & {\bf ja-en} & {\bf en-ko} & {\bf ko-en} & {\bf en-th} & {\bf th-en} & {\bf en-vi} & {\bf vi-en} \\ \hline
    \multicolumn{2}{c|}{\# Sentence Pairs}  & \multicolumn{2}{c|}{29.8M} & \multicolumn{2}{c|}{11.2M} & \multicolumn{2}{c|}{2.1M} & \multicolumn{2}{c|}{1.4M} & \multicolumn{2}{c|}{3.3M} & \multicolumn{2}{c}{3.5M}  \\ \hline
    \multirow{3}{*}{P@1} & Co-occurrence & 23.3 & 1.1 &  2.1 & 0.4 & 5.0 & 0.3 & 22.9 & 0.4 & 0.6 & 0.5 & 4.0 & 2.1 \\ 
                         & PMI           & 13.3 & 20.7 & 8.5 & 20.6 & 33.5 & 16.7 & 14.0 & 14.9 & 18.3 & 13.4 & 20.5 & 16.5 \\
                         & CPE           & {\bf 30.3} & {\bf 27.9} & {\bf 48.3} & {\bf 34.3} & {\bf 49.3} & {\bf 40.4} & {\bf 39.1} & {\bf 38.1} & {\bf 48.1} & {\bf 31.0} & {\bf 30.0} & {\bf 37.7} \\  \hline
    \multirow{3}{*}{P@5} & Co-occurrence & 46.9 & 35.2 & 50.5 & 27.1 & 30.7 & 29.1 & 36.6 & 26.9 & 55.6 & 24.4 & 39.3 & 28.3 \\
                         & PMI           & 57.0 & {\bf 61.6} & 78.7 & {\bf 65.3} & 64.0 & 60.5 & 48.8 & 57.7 & 64.5 & 52.7 & {\bf 50.1} & 60.4 \\
                         & CPE           & {\bf 58.1} & 50.5 & {\bf 80.9} & 60.1 & {\bf 66.8} & {\bf 66.4} & {\bf 54.9} & {\bf 60.0} & {\bf 69.3} & {\bf 53.1} & 48.9 & {\bf 62.2} \\                            
    \Xhline{1.1pt}
    \end{tabular}
    \caption{Precision (\%) on 2,000 word translations between six \emph{non-European} languages and English (source words randomly sampled from OpenSubtitles2018; gold labels taken from Google Translate).
    P@1 and P@5 denote the precision of top-1 and top-5 predictions, respectively.
    The ISO 639-1 language codes are used (ar: Arabic, zh: simplified Chinese, ja: Japanese, ko: Korean, th: Thai, vi: Vietnamese).}
    \label{tbl:results_noneuro}
\end{table*}

\begin{table*}[ht]
\centering
\begin{tabular}{l|l|l}
      \Xhline{1.1pt}
      \bf Language & \bf Python Tokenizer Module & \bf Reference \\
      \hline
      Arabic & \texttt{pyarabic.araby} & \cite{zerrouki2012pyarabic} \\
      Chinese (Simplified) & \texttt{Mykytea} & \cite{neubig2011pointwise}  \\
      Chinese (Traditional) & \texttt{jieba} & n/a  \\
      Japanese & \texttt{Mykytea} & \cite{neubig2011pointwise} \\
      Korean & \texttt{konlpy.tag.Mecab} & \cite{park2014konlpy} \\
      Thai & \texttt{pythainlp} & n/a  \\
      Vietnamese & \texttt{pyvi} & n/a  \\
      Others & \texttt{nltk.tokenize.TokTokTokenizer} & \cite{bird2009natural,dehdari2014neurophysiologically} \\
      \Xhline{1.1pt}
\end{tabular}
\caption{List of Python tokenizer modules used for each language.}
\label{tbl:tokenizers}
\end{table*}

\subsubsection{Evaluation on Non-European Languages}\label{sec:evaluation_noneuro}

Next, we compare the performance of co-occurrence, PMI, and CPE methods on language pairs between English and some of the major non-European languages: Arabic, simplified Chinese, Japanese, Korean, Thai, and Vietnamese.
As we detail in Section \ref{sec:tokenization}, these languages commonly require special word segmentation techniques.
Also, they typically have relatively smaller amounts of sentences paired with English, making it more challenging for the models to achieve high precision.

Unfortunately, we learned in our early experiments that the MUSE test set translations are far from being perfect for these non-European languages.
For example, in English-Vietnamese, we found that 80\% of the 1,500 word pairs in the test set had the same word twice as a pair (e.g. crimson-crimson, Suzuki-Suzuki, Randall-Randall).
Thus, for the non-European languages, we instead evaluate on translations using Google Translate\footnote{\scriptsize\url{https://translate.google.com/}}, a proprietary\footnote{We note that, because Google Translate is proprietary and not open-source, its results may change depending on the time of access. Our evaluations use Google Translate results accessed on July 19, 2019.} web software for machine translation.
To construct this test set, we first sample 2,000 words from the monolingual word distribution of that language pair's OpenSubtitles2018 parallel corpus. 
We use temperature-based smoothing ($T=1.25$) for the distribution to include more low-frequency words in the test set and also filter out words that include characters not from its alphabet (e.g., \emph{Charles} in Korean). 
Then, for each of the 2,000 sampled words, we retrieve ``common'' and ``uncommon'' translations\footnote{For word translations, Google Translate categorizes its translations to three categories: common, uncommon, and rare translations.} from Google Translate and treat them as ground truth labels.

The results are summarized in Table \ref{tbl:results_noneuro}.
Here, we see more evidence that the CPE method performs significantly better than both the co-occurrence and the PMI methods in top-1 precision as well as top-5 precision.
The performance gap tends to be larger both when the language's words are not whitespace-separated (e.g., Chinese and Japanese) and when there are a relatively small number of paired sentences (e.g., Korean and Thai).
Based on the results from Tables \ref{tbl:results_muse} and \ref{tbl:results_noneuro}, we employ the CPE method to produce the \emph{word2word} dataset.

\subsection{Word Segmentation}\label{sec:tokenization}

Since many of the 62 languages we consider are sensitive to word segmentation, we use language-specific tokenization tools when necessary. 
Specifically, we use publicly available tokenization packages for morphologically complex languages, i.e., Arabic \cite{attia2007arabic} and Korean, and languages in which words are not separated by spaces, i.e., Chinese, Japanese, Thai, and Vietnamese\footnote{Spaces in Vietnamese delimit syllables.}.
For all other languages, we use the tok-tok tokenizer \cite{dehdari2014neurophysiologically} implemented in NLTK \cite{bird2009natural}. 
Table \ref{tbl:tokenizers} summarizes the tokenization packages we used in the \emph{word2word} dataset and their references.

\section{The \emph{word2word} Python Interface}

As part of releasing the dataset and making it easily accessible and reproducible, we also introduce the \emph{word2word} Python package.
The open-source package provides an easy-to-use interface for both downloading and accessing bilingual lexicons for any of the 3,564 language pairs and building a custom bilingual lexicon on other language pairs for which there is a parallel corpus. 
Our source code is available on PyPi as \url{https://pypi.org/project/word2word/}. 

\subsection{Implementation}

The \emph{word2word} package is built entirely using Python 3.
The package includes scripts for downloading and pre-processing parallel corpora from OpenSubtitles2018, including word segmentation, and for computing the CPE scores for all available word tokens within each parallel corpus.
After processing, the package stores the bilingual lexicon as a Python \texttt{pickle} file, typically sized a few megabytes per language pair. 
The \texttt{pickle} file contains a Python dictionary that maps each source word to a list of top-10 word correspondences in $O(1)$ time. 
This allows bilingual lexicons to be portable and accessible.

\subsection{Usage}

The Python interface provides a simple API to download and access the \emph{word2word} dataset. 
As demonstrated in Figure \ref{fig:usage}, word translations for any query word can be retrieved as a list with a few lines of Python code.
\begin{figure}[htb]
    \centering
\begin{minted}[fontsize=\small]{python}
from word2word import Word2word

en2fr = Word2word('en', 'fr')
print(en2fr('apple'))
# ['pomme', 'pommes', 'pommier', 
#  'tartes', 'fleurs']
\end{minted}
    \caption{The \emph{word2word} Python interface for retrieving word translations.}
    \label{fig:usage}
\end{figure}

\subsection{Building a Custom Bilingual Lexicon}

The \emph{word2word} package also allows training a custom bilingual lexicon using a different parallel corpus. 
This can be useful in cases where there are larger and/or higher-quality parallel corpora available for the language pair of interest or when utilizing word translations for a particular domain (e.g., government, law, and medical). 
This process can also be done using a few lines of Python code, as demonstrated in Figure \ref{fig:custom}.
For an OpenSubtitles2018 corpus of a million parallel sentences, building a bilingual lexicon takes approximately 10 minutes using 8 CPUs.
\begin{figure}[htb]
    \centering
\begin{minted}[fontsize=\small]{python}
from word2word import Word2word

my_en2fr = Word2word.make(
    'en', 'fr', 'data/pubmed.en-fr'
)
# ...building...done!
print(my_en2fr('mitochondrial'))
# ['mitochondriale', 'mitochondriales', 
#  'mitochondrial', 'cytopathies', 
#  'mitochondriaux']
\end{minted}
    \caption{The \emph{word2word} Python interface for building a custom bilingual lexicon. Once built, the lexicon can be accessed in the same way as done in Figure \ref{fig:usage}.}
    \label{fig:custom}
\end{figure}

\section{Conclusion}

In this paper, we present the \emph{word2word} dataset, a publicly available collection of bilingual lexicons for 3,564 language pairs that are extracted from OpenSubtitles2018.
The bilingual lexicons have high coverage (up to hundreds of thousands words) for many language pairs and provide word translations of similar or better quality compared to those from a state-of-the-art embedding model.
We also release the \emph{word2word} Python package, with which the user can easily access the dataset or build a custom lexicon for different parallel corpora. 
We hope that the dataset and its Python interface can facilitate research on improving cross-lingual models, including machine translation models \cite{ramesh2018neural,gu2019pointer} and cross-lingual word embeddings \cite{conneau2017word,ruder2017survey}.



\section{Bibliographical References}
\label{main:ref}

\bibliographystyle{lrec}
\bibliography{word2word-lrec2020}

\begin{thebibliography}{}

\bibitem[\protect\citename{Artetxe \bgroup et al.\egroup
  }2018]{artetxe2018robust}
Artetxe, M., Labaka, G., and Agirre, E.
\newblock (2018).
\newblock A robust self-learning method for fully unsupervised cross-lingual
  mappings of word embeddings.
\newblock In {\em Proceedings of the 56th Annual Meeting of the Association for
  Computational Linguistics (Volume 1: Long Papers)}, pages 789--798,
  Melbourne, Australia, July. Association for Computational Linguistics.

\bibitem[\protect\citename{Artetxe \bgroup et al.\egroup
  }2019]{artetxe2019bilingual}
Artetxe, M., Labaka, G., and Agirre, E.
\newblock (2019).
\newblock Bilingual lexicon induction through unsupervised machine translation.
\newblock In {\em Proceedings of the 57th Annual Meeting of the Association for
  Computational Linguistics}, pages 5002--5007, Florence, Italy, July.
  Association for Computational Linguistics.

\bibitem[\protect\citename{Attia}2007]{attia2007arabic}
Attia, M.~A.
\newblock (2007).
\newblock Arabic tokenization system.
\newblock In {\em Proceedings of the 2007 workshop on computational approaches
  to semitic languages: Common issues and resources}, pages 65--72. Association
  for Computational Linguistics.

\bibitem[\protect\citename{Bird \bgroup et al.\egroup }2009]{bird2009natural}
Bird, S., Klein, E., and Loper, E.
\newblock (2009).
\newblock {\em Natural language processing with Python: analyzing text with the
  natural language toolkit}.
\newblock " O'Reilly Media, Inc.".

\bibitem[\protect\citename{Brown \bgroup et al.\egroup
  }1993]{brown1993mathematics}
Brown, P.~F., Pietra, V. J.~D., Pietra, S. A.~D., and Mercer, R.~L.
\newblock (1993).
\newblock The mathematics of statistical machine translation: Parameter
  estimation.
\newblock {\em Computational linguistics}, 19(2):263--311.

\bibitem[\protect\citename{Conneau \bgroup et al.\egroup
  }2017]{conneau2017word}
Conneau, A., Lample, G., Ranzato, M., Denoyer, L., and J{\'e}gou, H.
\newblock (2017).
\newblock Word translation without parallel data.
\newblock {\em arXiv preprint arXiv:1710.04087}.

\bibitem[\protect\citename{Dehdari}2014]{dehdari2014neurophysiologically}
Dehdari, J.
\newblock (2014).
\newblock {\em A Neurophysiologically-Inspired Statistical Language Model}.
\newblock {Ph.D.} thesis, The Ohio State University.

\bibitem[\protect\citename{Fung}1998]{fung1998statistical}
Fung, P.
\newblock (1998).
\newblock A statistical view on bilingual lexicon extraction: from parallel
  corpora to non-parallel corpora.
\newblock {\em Machine Translation and the Information Soup}, pages 1--17.

\bibitem[\protect\citename{Glava{\v{s}} \bgroup et al.\egroup
  }2019]{glavas2019properly}
Glava{\v{s}}, G., Litschko, R., Ruder, S., and Vuli{\'c}, I.
\newblock (2019).
\newblock How to (properly) evaluate cross-lingual word embeddings: On strong
  baselines, comparative analyses, and some misconceptions.
\newblock In {\em Proceedings of the 57th Annual Meeting of the Association for
  Computational Linguistics}, pages 710--721, Florence, Italy, July.
  Association for Computational Linguistics.

\bibitem[\protect\citename{Gouws \bgroup et al.\egroup }2015]{gouws2015bilbowa}
Gouws, S., Bengio, Y., and Corrado, G.
\newblock (2015).
\newblock Bilbowa: Fast bilingual distributed representations without word
  alignments.
\newblock In Francis Bach et~al., editors, {\em Proceedings of the 32nd
  International Conference on Machine Learning}, volume~37 of {\em Proceedings
  of Machine Learning Research}, pages 748--756, Lille, France, 07--09 Jul.
  PMLR.

\bibitem[\protect\citename{G{\=u} \bgroup et al.\egroup }2019]{gu2019pointer}
G{\=u}, J., Shavarani, H.~S., and Sarkar, A.
\newblock (2019).
\newblock Pointer-based fusion of bilingual lexicons into neural machine
  translation.
\newblock {\em arXiv preprint arXiv:1909.07907}.

\bibitem[\protect\citename{Koehn \bgroup et al.\egroup }2007]{koehn2007moses}
Koehn, P., Hoang, H., Birch, A., Callison-Burch, C., Federico, M., Bertoldi,
  N., Cowan, B., Shen, W., Moran, C., Zens, R., et~al.
\newblock (2007).
\newblock Moses: Open source toolkit for statistical machine translation.
\newblock In {\em Proceedings of the 45th annual meeting of the ACL on
  interactive poster and demonstration sessions}, pages 177--180. Association
  for Computational Linguistics.

\bibitem[\protect\citename{Levy and Goldberg}2014]{levy2014neural}
Levy, O. and Goldberg, Y.
\newblock (2014).
\newblock Neural word embedding as implicit matrix factorization.
\newblock In {\em Advances in Neural Information Processing Systems}, pages
  2177--2185.

\bibitem[\protect\citename{Levy \bgroup et al.\egroup }2017]{levy2017strong}
Levy, O., S{\o}gaard, A., and Goldberg, Y.
\newblock (2017).
\newblock A strong baseline for learning cross-lingual word embeddings from
  sentence alignments.
\newblock In {\em Proceedings of the 15th Conference of the European Chapter of
  the Association for Computational Linguistics: Volume 1, Long Papers},
  volume~1, pages 765--774.

\bibitem[\protect\citename{Lison \bgroup et al.\egroup
  }2018]{lison2018opensubtitles2018}
Lison, P., Tiedemann, J., and Kouylekov, M.
\newblock (2018).
\newblock Opensubtitles2018: Statistical rescoring of sentence alignments in
  large, noisy parallel corpora.
\newblock In {\em Proceedings of the Eleventh International Conference on
  Language Resources and Evaluation (LREC 2018)}.

\bibitem[\protect\citename{Liu \bgroup et al.\egroup }2013]{liu2013topic}
Liu, X., Duh, K., and Matsumoto, Y.
\newblock (2013).
\newblock Topic models + word alignment = a flexible framework for extracting
  bilingual dictionary from comparable corpus.
\newblock In {\em Proceedings of the Seventeenth Conference on Computational
  Natural Language Learning}, pages 212--221, Sofia, Bulgaria, August.
  Association for Computational Linguistics.

\bibitem[\protect\citename{Mikolov \bgroup et al.\egroup
  }2013a]{mikolov2013exploiting}
Mikolov, T., Le, Q.~V., and Sutskever, I.
\newblock (2013a).
\newblock Exploiting similarities among languages for machine translation.
\newblock {\em arXiv preprint arXiv:1309.4168}.

\bibitem[\protect\citename{Mikolov \bgroup et al.\egroup
  }2013b]{mikolov2013distributed}
Mikolov, T., Sutskever, I., Chen, K., Corrado, G.~S., and Dean, J.
\newblock (2013b).
\newblock Distributed representations of words and phrases and their
  compositionality.
\newblock In {\em Advances in neural information processing systems}, pages
  3111--3119.

\bibitem[\protect\citename{Neubig \bgroup et al.\egroup
  }2011]{neubig2011pointwise}
Neubig, G., Nakata, Y., and Mori, S.
\newblock (2011).
\newblock Pointwise prediction for robust, adaptable japanese morphological
  analysis.
\newblock In {\em Proceedings of the 49th Annual Meeting of the Association for
  Computational Linguistics: Human Language Technologies: short papers-Volume
  2}, pages 529--533. Association for Computational Linguistics.

\bibitem[\protect\citename{Park and Cho}2014]{park2014konlpy}
Park, E.~L. and Cho, S.
\newblock (2014).
\newblock Konlpy: Korean natural language processing in python.
\newblock In {\em Proceedings of the 26th Annual Conference on Human \&
  Cognitive Language Technology}, Chuncheon, Korea, October.

\bibitem[\protect\citename{Ramesh and Sankaranarayanan}2018]{ramesh2018neural}
Ramesh, S.~H. and Sankaranarayanan, K.~P.
\newblock (2018).
\newblock Neural machine translation for low resource languages using bilingual
  lexicon induced from comparable corpora.
\newblock In {\em Proceedings of the 2018 Conference of the North {A}merican
  Chapter of the Association for Computational Linguistics: Student Research
  Workshop}, pages 112--119, New Orleans, Louisiana, USA, June. Association for
  Computational Linguistics.

\bibitem[\protect\citename{Ruder \bgroup et al.\egroup }2017]{ruder2017survey}
Ruder, S., Vuli\'c, I., and S{\o}gaard, A.
\newblock (2017).
\newblock A survey of cross-lingual embedding models.
\newblock {\em arXiv preprint arXiv:1706.04902}.

\bibitem[\protect\citename{Somers}2001]{somers2001bilingual}
Somers, H.
\newblock (2001).
\newblock Bilingual parallel corpora and language engineering.
\newblock In {\em Proc. Anglo-Indian Workshop" Language Engineering for
  South-Asian languages}.

\bibitem[\protect\citename{Vogel \bgroup et al.\egroup }1996]{vogel1996hmm}
Vogel, S., Ney, H., and Tillmann, C.
\newblock (1996).
\newblock Hmm-based word alignment in statistical translation.
\newblock In {\em Proceedings of the 16th conference on Computational
  linguistics-Volume 2}, pages 836--841. Association for Computational
  Linguistics.

\bibitem[\protect\citename{Vuli{\'c} and Moens}2013]{vulic2013cross}
Vuli{\'c}, I. and Moens, M.-F.
\newblock (2013).
\newblock Cross-lingual semantic similarity of words as the similarity of their
  semantic word responses.
\newblock In {\em Proceedings of the 2013 Conference of the North {A}merican
  Chapter of the Association for Computational Linguistics: Human Language
  Technologies}, pages 106--116, Atlanta, Georgia, June. Association for
  Computational Linguistics.

\bibitem[\protect\citename{Zerrouki}2010]{zerrouki2012pyarabic}
Zerrouki, T.
\newblock (2010).
\newblock pyarabic, an arabic language library for python.

\end{thebibliography}


\appendix

\clearpage

\setlength\tabcolsep{2pt}
\begin{table*}[hbt!]
    \centering
    \small
    \begin{tabular}{c | c | c c c c c | c c c c c}
        \Xhline{1.1pt}
        {\bf English} & {\bf Methods} & \multicolumn{5}{c}{{\bf Top-5 Translations in Spanish}} & \multicolumn{5}{|c}{{\bf Top-5 Translations in Simplified Chinese}} \\ \hline
        \multirow{3}{*}{its}  &  Co-occurrence  & de  &   la  &   que  &   el   &   y  & \zh{的}	& \zh{\bf 它}	& \zh{了}	& \zh{是}	& \zh{我}  \\
         & PMI & propia & {\bf sus}  &  {\bf su}  &    tierra   &  poder & \zh{\bf 它}  &	\zh{政府}   &	\zh{国家}   &	\zh{失去}   &	\zh{由}  \\
         & CPE & {\bf su}  & {\bf sus}  &  propia  &  tierra &  cada & \zh{\bf 它}	 & \zh{将}	& \zh{自己} & 	\zh{国家} &	\zh{中} \\ \hline
        \multirow{3}{*}{good} & Co-occurrence & que & de & no & {\bf bien} & es & \zh{\bf 好} & \zh{的} & \zh{你} & \zh{我} & \zh{很} \\
        & PMI & {\bf buenas} & noches & {\bf buenos} & {\bf buena} & {\bf buen} & \zh{祝你好运} & \zh{晚安} & \zh{好消息} & \zh{早上好} & \zh{早安} \\
        & CPE & {\bf bien} & {\bf buena} & {\bf buenas} & {\bf buen} & {\bf bueno} & \zh{\bf 好} & \zh{很} & \zh{\bf 不错} & \zh{晚安} & \zh{早上好} \\ \hline
        \multirow{3}{*}{mouth}  &  Co-occurrence  & la & {\bf boca} & de & que & no & \zh{的} &	\zh{你} & \zh{我} &	\zh{\bf 嘴} & \zh{了} \\
         & PMI & {\bf boca} &	cerrada & 	pico &	mant\'en &	abre & 
        \zh{张开嘴}	& \zh{嘴里} & \zh{大嘴巴} & \zh{张嘴} & \zh{\bf 嘴巴} \\
         & CPE & {\bf boca} &	cerrada & 	abre	& palabras &	labios & \zh{\bf 嘴} &	\zh{嘴里} & 	\zh{\bf 嘴巴} &	\zh{闭上} &	\zh{闭嘴}  \\ \hline
        \multirow{3}{*}{library}  &  Co-occurrence  & la & {\bf biblioteca} & de & en & que & \zh{\bf 图书馆}	&  \zh{的}	&   \zh{我}	&   \zh{在}	&   \zh{你} \\
         & PMI & solarizaci\'on & {\bf biblioteca} & solt\'andola & library &   librer\'ia & \zh{莊英國} &	\zh{\bf 图书馆}	&   \traditionalchinese{\bf 圖書館}  &	\zh{藏书室} &	\zh{书房}\\
         & CPE & {\bf biblioteca} & la & librer\'ia  &  p\'ublica &  tarjetas & \zh{\bf 图书馆} &	\zh{书房}	&  \zh{里}	&  \traditionalchinese{\bf 圖書館}	&  \zh{去}  \\
         \Xhline{1.1pt}
    \end{tabular}
    \caption{Selected \emph{word2word} translations of English words into Spanish and simplified Chinese. Top-5 predictions are listed in the decreasing order of the model's scores. Boldfaced target words indicate correct translations.}
    \label{tbl:examples_comparison}
\end{table*}

\section{Sample Translations from Different Extraction Methods}\label{app:comparison}

In Table \ref{tbl:examples_comparison}, we compare the BLE methods described in Section \ref{sec:ble} from illustrative examples of their extracted bilingual lexicons for English to Spanish and English to Simplified Chinese.
These examples show that the CPE approach provides the correct correspondence as its top-1 translation in both languages, while the PMI approach seems to excessively favor rarer words among the co-occurrences.
As illustrated in the English-Chinese example, this can be particularly problematic with languages such as Chinese, where word segmentation is highly nontrivial. 
The co-occurrence method prefers stop words that are frequent over the entire document, rather than the corresponding words.

\subsection{Co-occurrences}

The baseline co-occurrence model performs poorly in both experiments (Tables \ref{tbl:results_muse} and \ref{tbl:results_noneuro}).
As exemplified in Table \ref{tbl:examples_comparison}, we find that the top-5 predictions in many cases are primarily stop words, such as \emph{la} (the), \emph{de} (of), and \emph{que} (that) in Spanish and \zh{的} (of), \zh{你} (you), and \zh{我} (I, me) in Chinese, because they frequently occur in any sentence, regardless of context. 

\subsection{Comparing PMI and CPE}

Comparing translations using PMI and CPE, we find in Table \ref{tbl:examples_comparison} that PMI favors less frequent words excessively. 
This results in two kinds of error cases: (a) when PMI overemphasizes rare words in the target vocabulary, e.g. \emph{solarizaci\'on} for \emph{library} in en-es, and (b) when PMI misses correct words in the target language that are relatively frequently used, e.g. \emph{bien} for \emph{good} in en-es. 
Another consequence is that PMI prefers less common variants of the same word, in particular conjugations and past/future tenses as well as typos, when two forms of the same word have comparable counts (e.g. \emph{obligados} preferred over \emph{obligado} in Spanish for the English \emph{obliged}).

Because of the second reason, we also find that \emph{word2word} tends to be more robust to tokenization issues, which are common in non-whitespace-separated languages like Chinese. 
For example, since the tokenizer failed to separate \zh{张开嘴} (open mouth), which in general occurs far less frequently than \zh{嘴} (mouth), PMI favors \zh{张开嘴} over the more frequent \zh{嘴} as its first choice.

\section{Full Dataset Statistics}\label{app:counts}

In Table \ref{tbl:counts}, we list the sizes of all 3,564 bilingual lexicons in the \emph{word2word} dataset. 
By size, we refer to the number of source words for which translations exist. 
For each source word, we extract up to 10 (9+ on average) most likely translations according to the CPE method described in \ref{sec:cpe}

\begin{sidewaystable*}
\centering
\tiny
\setlength\tabcolsep{0pt}
\begin{tabular}{c|c|c|c|c|c|c|c|c|c|c|c|c|c|c|c|c|c|c|c|c|c|c|c|c|c|c|c|c|c|c|c|c|c|c|c|c|c|c|c|c|c|c|c|c|c|c|c|c|c|c|c|c|c|c|c|c|c|c|c|c|c|c}
\Xhline{1.1pt}
          & \bf af & \bf ar & \bf bg & \bf bn & \bf br & \bf bs & \bf ca & \bf cs & \bf da & \bf de & \bf el & \bf en & \bf eo & \bf es & \bf et & \bf eu & \bf fa & \bf fi & \bf fr & \bf gl & \bf he & \bf hi & \bf hr & \bf hu & \bf hy & \bf id & \bf is & \bf it & \bf ja & \bf ka & \bf kk & \bf ko & \bf lt & \bf lv & \bf mk & \bf ml & \bf ms & \bf nl & \bf no & \bf pl & \bf pt & \bf pt\_br & \bf ro & \bf ru & \bf si & \bf sk & \bf sl & \bf sq & \bf sr & \bf sv & \bf ta & \bf te & \bf th & \bf tl & \bf tr & \bf uk & \bf ur & \bf vi & \bf ze\_en & \bf ze\_zh & \bf zh\_cn & \bf zh\_tw \\
\end{tabular}
\caption{Bilingual lexicon counts for the entire \emph{word2word} dataset. Each (row, column) entry is the number of words in the (row $\rightarrow$ column) bilingual lexicon. See \url{http://opus.nlpl.eu/OpenSubtitles2018.php} for language codes and original data sizes.}
\label{tbl:counts}
\end{sidewaystable*}

\end{document}